\documentclass[letterpaper]{article} 
\usepackage{aaai2026}  
\usepackage{times}  
\usepackage{helvet}  
\usepackage{courier}  
\usepackage[hyphens]{url}  
    \usepackage{graphicx} 
\usepackage{natbib}  
\usepackage{caption} 
\usepackage{algorithm}
\usepackage{algorithmic}
\usepackage{amsmath} 
\usepackage{booktabs}
\usepackage{multirow}
\usepackage{amssymb}
\usepackage{url}
\usepackage{newfloat}
\usepackage{listings}
\DeclareCaptionStyle{ruled}{labelfont=normalfont,labelsep=colon,strut=off} 
\floatstyle{ruled}
\newfloat{listing}{tb}{lst}{}
\floatname{listing}{Listing}
\title{uCLIP: Parameter-Efficient Multilingual Extension of Vision-Language Models with Unpaired Data}

\author{
    Dahyun Chung\equalcontrib\textsuperscript{\rm 1},
    Donghyun Shin\equalcontrib\textsuperscript{\rm 1},
    Yujin Sung\equalcontrib\textsuperscript{\rm 1},
    Seunggi Moon\equalcontrib\textsuperscript{\rm 1},\\
    Jinwoo Jeon\textsuperscript{\rm 1},
    Byung-Jun Lee\textsuperscript{\rm 1}
}

\affiliations{
    \textsuperscript{\rm 1}Korea University\\
    \{dahyun1016, dawnme, sung031128, moon44432, kevin04087, byungjunlee\}@korea.ac.kr
}

\begin{document}

\maketitle

\begin{abstract}
Contrastive Language–Image Pre-training (CLIP) has demonstrated strong generalization across a wide range of visual tasks by leveraging large-scale English–image pairs. However, its extension to low-resource languages remains limited due to the scarcity of high-quality multilingual image–text data. Existing multilingual vision–language models exhibit consistently low retrieval performance in underrepresented languages—including Czech, Finnish, Croatian, Hungarian, Romanian—on the Crossmodal-3600 (XM3600) benchmark. To address this, we propose a lightweight and data-efficient framework for multilingual vision–language alignment. Our approach requires no image–text pairs or text-text pairs and freezes both the pretrained image encoder and multilingual text encoder during training. Only a compact 1.7M-parameter projection module is trained, using a contrastive loss over English representations as semantic anchors. This minimal training setup enables robust multilingual alignment even for languages with limited supervision. Extensive evaluation across multiple multilingual retrieval benchmarks confirms the effectiveness of our method, showing significant gains in five underrepresented languages where existing models typically underperform. These findings highlight the effectiveness of our pivot-based, parameter-efficient alignment strategy for inclusive multimodal learning. Our code is
available at \url{https://dinyudin203.github.io/uCLIP-project/}
\end{abstract}


\section{Introduction}
\begin{figure*}[t]
  \centering
  \includegraphics[width=\textwidth]{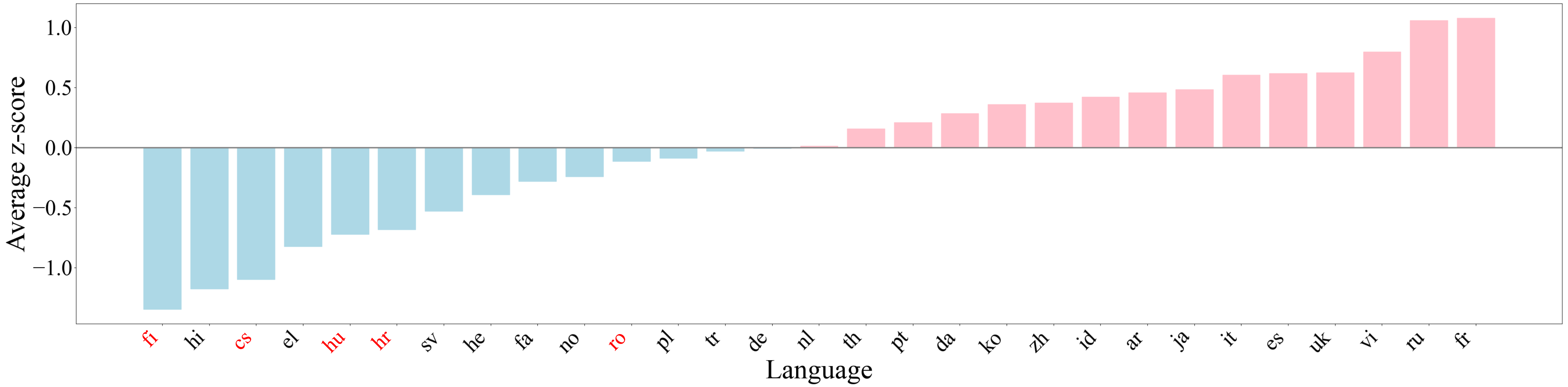}
\caption{
\textbf{Average z-score of Recall@10 across languages.}
We evaluate multilingual VLM performance on the XM3600 benchmark using Recall@10 from four baseline models: AltCLIP-18, SigLIP2, NLLB-CLIP, and M-CLIP. For each model, we compute the z-score per language, indicating how much its Recall@10 deviates from the model-wise mean. The final score is the average z-score across models. Languages highlighted in red represent the five low-resource languages we target (cs, fi, hr,
hu, ro). Unsupported languages by our multilingual text encoder (e.g., bn, fil, mi, quz, sw, te) are excluded from our evaluation.
}
  \label{fig:average_zscore}
\end{figure*}

Contrastive vision–language models such as CLIP~\citep{radford2021learning} have demonstrated strong generalization ability by training on large-scale English–image parallel datasets. However, their effectiveness does not easily transfer to low-resource languages, primarily due to the lack of high-quality multilingual parallel text–image datasets. This data imbalance limits the accessibility and fairness of vision–language technologies in multilingual settings, despite their increasing demand in areas like cross-lingual retrieval~\citep{Litschko2018retrieval}, content recommendation~\citep{Liu2021recommendation}, and digital media understanding~\citep{liu2022review}.
A common workaround involves translation-based pipelines (e.g., translating text before applying CLIP), which bypass the need for multilingual supervision. However, these approaches not only introduce additional latency but also often suffer from semantic drift on ambiguous or culturally specific expressions—particularly in low-resource languages.~\cite{qiu2022multilingual} Multilingual vision–language models (VLMs) have emerged as an alternative or complement to translation-based pipelines, aligning multilingual text with images directly. However, even state-of-the-art multilingual VLMs continue to underperform on underrepresented languages, as shown in Figure~\ref{fig:average_zscore} and Supplementary Material.
\begin{figure}[t]
  \centering
  \includegraphics[width=\linewidth]{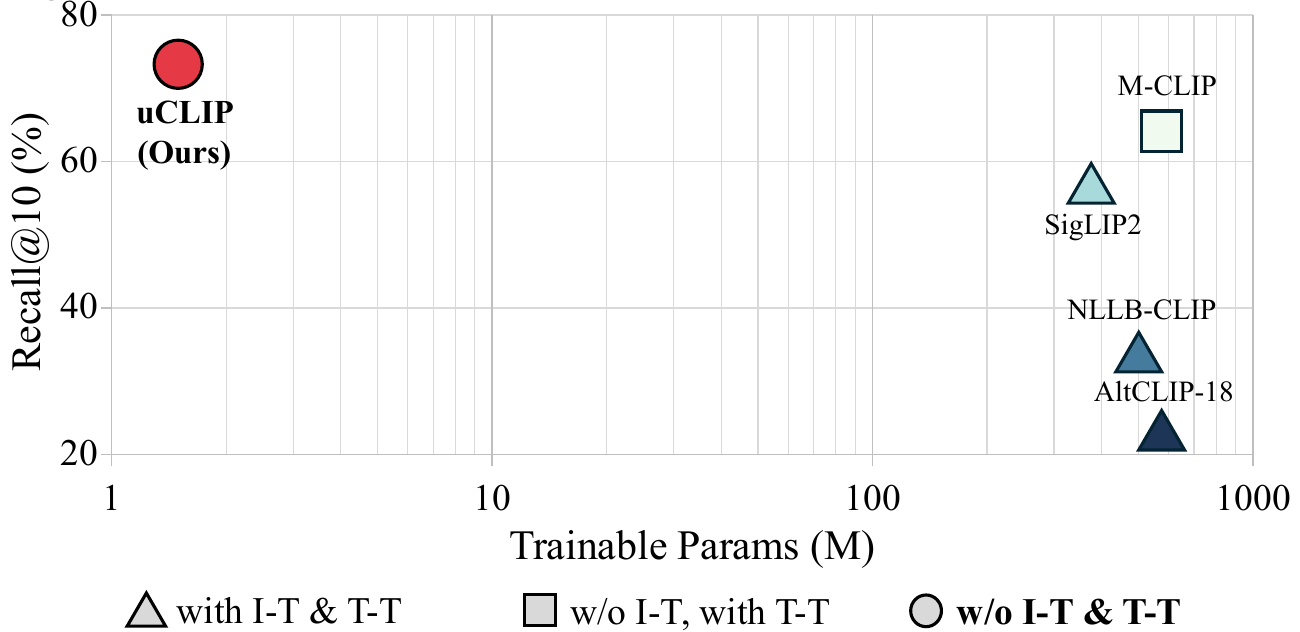}
\caption{\textbf{Performance Comparison.} 
We compare models by plotting the average image-to-text Recall@10 across five underrepresented languages in the XM3600 benchmark against the number of trainable parameters (in millions). 
Each marker shape indicates the type of supervision used during training: 
\textit{circle} for models trained without image-text (multilingual or English) pairs (I-T) or multilingual-English text pairs (T-T), 
\textit{square} for those trained with T-T pairs only, and 
\textit{triangle} for models using both I-T and T-T pairs.
Despite having only 1.7M parameters and no paired supervision, \textit{uCLIP} achieves the highest average Recall@10, outperforming all baselines.}

  \label{fig:teaser}
\end{figure}
Existing multilingual VLMs largely follow one of the two approaches. The first strategy adopts a frozen image encoder with a fine-tuned multilingual text encoder, e.g., AltCLIP~\citep{chen2022altclip}, M-CLIP~\citep{carlsson2022multilingualclip}, and NLLB-CLIP~\citep{visheratin2023nllb}. Such methods preserve CLIP’s image encoder and substitute its English text encoder with a multilingual model (e.g., XLM-R, mBERT, NLLB). Training focuses solely on aligning multilingual text with the fixed visual representation, typically via translated or synthetic data with contrastive or distillation-based objectives. While this design retains the benefits of CLIP’s visual representations, it typically relies on image–text (multilingual or English) pairs (I-T) or multilingual–English text pairs (T-T) to align the new multilingual encoder with the visual space. For example, M-CLIP performs knowledge distillation from the CLIP text encoder to mBERT, requiring large-scale paired T-T corpora. AltCLIP aligns XLM-R to the CLIP text encoder using T-T pairs, and further fine-tunes it with I-T pairs to match the frozen image encoder. NLLB-CLIP fine-tunes the NLLB encoder with translated captions using contrastive loss against the CLIP image encoder, again depending on I-T paired supervision.

The second one involves joint training of both image and text encoders, e.g., SigLIP2~\citep{zhai2023siglip}. These models are trained from scratch or heavily fine-tuned using massive multilingual datasets (e.g., WebLI) and employ novel objectives such as sigmoid contrastive loss to encourage dense alignment. This design facilitates a more flexible multilingual-visual representation space and theoretically better captures diverse semantic structures across languages. However, it comes at the cost of massive computation, requires extensive resources, and struggles to generalize across languages with limited training data.

To address this limitation, we propose \textit{uCLIP}, an efficient multilingual vision–language model that eliminates the need for paired supervision. Rather than relying on image–text or multilingual–English text pairs, uCLIP adopts a pivot-based strategy that uses English as an implicit bridge for multilingual alignment. This design trains only a lightweight projection module, enabling training efficiency and strong zero-shot performance.

Specifically, our method leverages the robust English–image alignment in CLIP and transfers this to other languages via a frozen multilingual text encoder. The architecture comprises a frozen CLIP image encoder, a frozen multilingual text encoder, and a compact projection head with just \textbf{1.7M trainable parameters}. This design preserves the strong performance of CLIP achieved through large-scale pretraining, without requiring paired I–T or T–T data, or retraining large encoders. As a result, \textit{uCLIP} achieves effective multilingual retrieval and classification, improving performance in underrepresented languages while significantly reducing training cost. Our contributions are as follows.
\begin{itemize}
    \item We propose \textit{uCLIP}, a novel multilingual vision-language alignment framework that eliminates the need for image–text (I–T) pairs—whether multilingual or English—and multilingual–English text (T–T) pairs during training, by leveraging English as a universal semantic anchor.
    \item  Our method trains only a lightweight projection module, enabling training-efficient alignment with over 99\% fewer trainable parameters than prior baselines.
    \item \textit{uCLIP} demonstrates strong zero-shot performance on multiple text-image retrieval and classification benchmarks, while also achieving significantly lower inference latency, demonstrating both effectiveness and efficiency.
\end{itemize}

Given space limitations and the wide range of existing studies, we provide detailed discussions of related work in the Supplementary Material.
\section{Proposed Method}
\begin{figure*}[t]
  \centering
  \includegraphics[width=\textwidth]{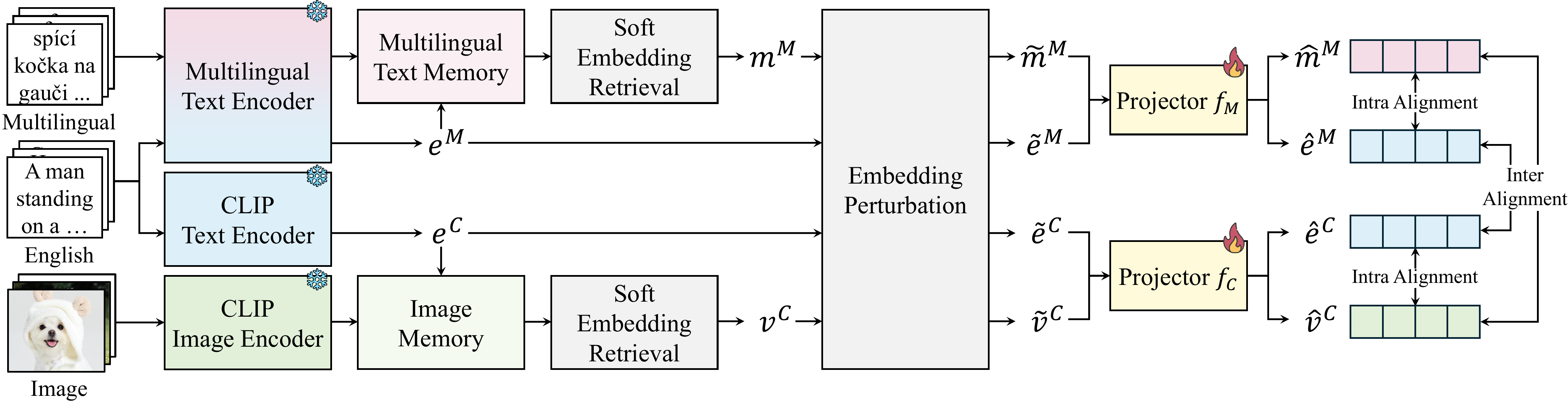}
\caption{
\textbf{Overall architecture.} We propose a lightweight alignment framework that bridges multilingual text and image embeddings via English, without requiring paired I-T and T-T data or encoder finetuning. \textit{uCLIP} employs frozen encoders along with compact projection heads to map inputs into a shared embedding space. At inference time, only multilingual text encoder, image encoder and projectors are used. The model directly encodes multilingual text and image inputs using the frozen encoders, followed by projection into the shared space.}
  \label{fig:architecture}
\end{figure*}

In this section, we present our proposed method for aligning multilingual text and images using only projectors, with English serving as the semantic pivot. The approach constructs a shared embedding space through memory-based retrieval and optimization of an alignment loss. An overview of the method is shown in Figure~\ref{fig:architecture}, with a detailed explanation provided below.

\subsection{Soft Embedding Retrieval}
To enable multilingual alignment without requiring paired I-T, T-T datasets, we use English queries as semantic anchors to retrieve pseudo-aligned representations from memory banks.
Given an English query \( e_i \), we extract two embeddings: \( e^C_i \) from the CLIP text encoder and \( e^M_i \) from a multilingual text encoder. We also construct two memory banks: image memory \( V = \{v_1, \dots, v_N\} \) encoded by the CLIP image encoder, and multilingual text memory \( M = \{m_1, \dots, m_K\} \) encoded by the same multilingual text encoder shared across target languages. Details on memory bank sizes, sampling strategies, and data sources are provided in the Supplementary Material.

To retrieve semantically aligned features, we compute softmax-weighted averages in each memory bank based on cosine similarity. The image representation aligned with the English query is retrieved as:

\begin{equation}
v^C_i = \sum_{k=1}^N \frac{\exp(\text{sim}(e^C_i, v_k)/\tau)}{\sum_{j=1}^N \exp(\text{sim}(e^C_i, v_j)/\tau)} \cdot v_k
\label{eq:v_i^C}
\end{equation}
Likewise, the multilingual text embedding aligned with the same query is:
\begin{equation}
m^M_i = \sum_{k=1}^K \frac{\exp(\text{sim}(e^M_i, m_k)/\tau)}{\sum_{j=1}^K \exp(\text{sim}(e^M_i, m_j)/\tau)} \cdot m_k
\label{eq:m_i^M}
\end{equation}
These soft-retrieved representations \( v^C_i\) and \( m^M_i\) serve as pseudo-aligned features that bridge modality through the shared semantics of the English query.
\subsection{Embedding Perturbation}
Multilingual text encoders often produce unstable or noisy representations, especially for low-resource languages. To enhance robustness against encoding biases, we add Gaussian noise to the four embeddings \( e^C, e^M, v^C, m^M \), and project them onto the unit hypersphere:
\begin{alignat}{2}
\tilde{e}_i^C &= \text{Norm}(e_i^C + \epsilon_1), \quad &
\tilde{e}_i^M &= \text{Norm}(e_i^M + \epsilon_2) \label{eq:noise1} \\
\tilde{v}_i^C &= \text{Norm}(v_i^C + \epsilon_3), \quad &
\tilde{m}_i^M &= \text{Norm}(m_i^M + \epsilon_4) \label{eq:noise2}
\end{alignat}
where \( \epsilon_1, \epsilon_2, \epsilon_3, \epsilon_4 \sim \mathcal{N}(0, \sigma^2 I) \). 
This perturbation encourages the model to align local neighbors of embeddings rather than relying on exact vectors.
As a result, it promotes local semantic smoothness and enhances robustness across modalities. 
\subsection{Inter Alignment}
For each English query \( i \), we first obtain four perturbed embeddings: \( \tilde{e}_i^C \) and \( \tilde{e}_i^M \) denote the representations of the same English query obtained from the CLIP text encoder and the multilingual text encoder, respectively; \( \tilde{v}_i^C \) and \( \tilde{m}_i^M \) are the image and multilingual text features retrieved from the memory banks using these text embeddings.
To project these embeddings into a shared embedding space, we group them by their source modality. The CLIP text embedding (\(\tilde{e}_i^C\)) and the retrieved image feature (\(\tilde{v}_i^C\)) both originate from the image-text modality space associated with CLIP. Therefore, they are projected by a shared clip projection head, \(f_C\). Similarly, the multilingual text embedding (\(\tilde{e}_i^M\)) and the retrieved multilingual text feature(\(\tilde{m}_i^M\)) belong to the multilingual text space and are thus projected by a shared multilingual projection head, \(f_M\). 
We use two different projection heads to accommodate statistical differences between independently pretrained encoders, which improves alignment stability across modalities.
This process is formulated as follows:
\begin{alignat}{2}
\hat{e}_i^C &= f_C(\tilde{e}_i^C), \quad &
\hat{e}_i^M &= f_M(\tilde{e}_i^M) \nonumber \\
\hat{v}_i^C &= f_C(\tilde{v}_i^C), \quad &
\hat{m}_i^M &= f_M(\tilde{m}_i^M) \nonumber
\end{alignat}
The pair \( (\hat{e}_i^C, \hat{e}_i^M) \), which represents the same English query encoded by two distinct encoders, provides a strong supervision signal to align textual semantics across encoders. On the other hand, the pair \( (\hat{v}_i^C, \hat{m}_i^M) \), which associates retrieved image and multilingual text embeddings, serves as a pseudo-aligned cross-modal pair.

To enforce alignment between the embedding pairs from different encoders—either \( (\hat{e}_i^C, \hat{e}_i^M) \) and \( (\hat{v}_i^C, \hat{m}_i^M) \)—we apply contrastive losses to both pairs.
The InfoNCE loss between query embedding $q$ and key embedding $k$ is defined as :
\begin{align}
\ell(q, k) = -\frac{1}{B} \sum_{i=1}^{B} \Bigg[ \log \frac{\exp(\text{sim}(q_i, k_i)/\tau)}{\sum_{j=1}^{B} \exp(\text{sim}(q_i, k_j)/\tau)} \Bigg]
\end{align}

The similarity function \(\text{sim}(\cdot, \cdot)\) denotes cosine similarity between two embeddings. 
Since softmax normalization is asymmetric, we both compute $\ell(\hat{e}^C, \hat{e}^M)$, $\ell(\hat{e}^M, \hat{e}^C)$ and take the average to ensure robust alignment between representations. The alignment loss between English text embeddings (\(\hat{e}^C, \hat{e}^M\)) is defined as:

\begin{align}
\mathcal{L}_{\text{text}} := \frac{1}{2}\bigg[\ell(\hat{e}^C, \hat{e}^M) + \ell(\hat{e}^M, \hat{e}^C)\bigg] 
\nonumber 
\end{align} 

The same formulation is used for the pseudo-aligned image and multilingual text embeddings (\(\hat{v}^C, \hat{m}^M\)):
\begin{align}
\mathcal{L}_{\text{pseudo}} := \frac{1}{2}\bigg[\ell(\hat{v}^C, \hat{m}^M) + \ell(\hat{m}^M, \hat{v}^C)\bigg]  
\nonumber
\end{align} 
The total inter-alignment loss is defined as the sum of these two:
\begin{equation*}
\mathcal{L}_{\text{inter}} := \mathcal{L}_{\text{text}} + \mathcal{L}_{\text{pseudo}}
\end{equation*}
The alignment between \( \hat{e}_i^C \) and \( \hat{e}_i^M \), derived from the same English query, acts as a strong supervisory signal. This supervision helps guide the learning of the pseudo-aligned pair \( (\hat{v}_i^C, \hat{m}_i^M) \), facilitating robust cross-modal and cross-lingual representation learning. Notably, this design enables \textit{uCLIP} to be trained without requiring any paired I-T and T-T data, as the English embedding alignment provides transferable supervision across modalities and languages. By anchoring alignment on high-resource English queries, \textit{uCLIP} transfers robust supervision to underrepresented languages.

\subsection{Intra Alignment}
Semantically aligned embeddings—such as \( \hat{e}_i^C \) and \( \hat{v}_i^C \), or \( \hat{e}_i^M \) and \( \hat{m}_i^M \)—often lie in disjoint regions due to modality gaps, especially when pretrained encoders are used without joint tuning. To address this, we introduce an intra-modality loss that pulls such embeddings closer, encouraging a cohesive and isotropic shared space that complements the inter-modality alignment.
Inspired by \citet{liang2022mind}, we remove the repulsive term from contrastive loss and retain only the attractive component.
In contrastive learning, the standard InfoNCE loss can be decomposed as:
\begin{equation*}
\begin{aligned}
    & -\log \frac{\exp(\mathrm{sim}(x_i, z_i)/\tau)}{\sum_{j=1}^{N} \exp(\mathrm{sim}(x_i, z_j)/\tau)} \\
    & \quad = -\frac{1}{\tau} \mathrm{sim}(x_i, z_i)+ \log \sum_{j=1}^{N} \exp\left( \frac{\mathrm{sim}(x_i, z_j)}{\tau} \right)
\end{aligned}
\end{equation*}
where the first term attracts positive pairs and the second term repels negatives. Since all embeddings in our model are $\ell_2$-normalized, cosine similarity and squared Euclidean distance are functionally equivalent:
\begin{equation*}
    \| x - y \|_2^2  = 2(1 - x^\top y) = 2(1 - \mathrm{sim}(x, y))
\end{equation*}
Thus, maximizing cosine similarity is equivalent to minimizing Euclidean distance in the normalized space. We leverage this equivalence and define the intra-alignment loss using the attractive term only:
\begin{equation*}
    \mathcal{L}_{\text{intra}} := \frac{1}{2B} \sum_{i=1}^{B} \left(
    \left\| \hat{e}_i^C - \hat{v}_i^C \right\|_2^2 + 
    \left\| \hat{e}_i^M - \hat{m}_i^M \right\|_2^2
    \right)
\end{equation*}
This encourages angular proximity and closes modality gaps within both CLIP and multilingual spaces. The intra-alignment complements inter-alignment, enabling robust text–image representation alignment via English intermediaries.

In summary, our final loss combines inter- and intra-alignment objectives: 
\begin{equation*}
    \mathcal{L} := \mathcal{L}_{\text{inter}} + \lambda \mathcal{L}_{\text{intra}}. 
\end{equation*}
The inter-alignment loss encourages cross-modal and cross-lingual consistency via English as a semantic pivot. This formulation enables \textit{uCLIP} to learn a unified multilingual text-image space without relying on paired data in target languages.

\begin{table*}[t]
    \centering
    \setlength{\tabcolsep}{7pt}
    \footnotesize
        \begin{tabular}{l | c  c  c  c c  c  c  c  c  c | c  c}
        \toprule
        \multirow{2}{*}{Method} &
        \multicolumn{2}{c}{cs} & \multicolumn{2}{c}{fi} & \multicolumn{2}{c}{hr} &
        \multicolumn{2}{c}{hu} & \multicolumn{2}{c}{ro} & \multicolumn{2}{|c}{\textbf{Avg.}} \\
        & I$\rightarrow$T & T$\rightarrow$I & I$\rightarrow$T & T$\rightarrow$I &
          I$\rightarrow$T & T$\rightarrow$I & I$\rightarrow$T & T$\rightarrow$I &
          I$\rightarrow$T & T$\rightarrow$I & I$\rightarrow$T & T$\rightarrow$I \\
        \midrule
        
        \multicolumn{13}{c}{\textit{MSCOCO (R@10)}} \\
        \midrule
        AltCLIP-18        & 18.2 & 16.0 &  7.6 &  4.6 & 14.6 & 11.8 & 11.5 &  9.6 & 21.6 & 18.4 & 14.7 & 12.1 \\
        SigLIP2           & \underline{51.4} & 46.7 & 30.6 & 24.6 & 40.0 & 35.2 & \underline{48.4} & 43.3 & \underline{53.3} & 47.2 & \underline{44.7} & 39.4 \\
        NLLB-CLIP         & 28.6 & 31.3 & 10.1 &  8.7 & 23.0 & 25.1 & 22.4 & 23.2 & 40.1 & 42.9 & 24.8 & 26.2 \\
        M-CLIP            & 40.1 & \textbf{54.8} & \underline{31.8} & \underline{48.1} & \underline{42.7} & \textbf{54.8} & 42.2 & \textbf{53.5} & 39.5 & \textbf{56.3} & 39.3 & \textbf{53.5}\\
        \textbf{uCLIP (Ours)} & \textbf{54.9} & \underline{54.2} & \textbf{50.0} & \textbf{49.9} & \textbf{53.6} & \underline{54.3} & \textbf{52.2} & \underline{53.2} & \textbf{55.4} & \underline{54.8} & \textbf{53.2}& \underline{53.3}\\
        \midrule
        
        \multicolumn{13}{c}{\textit{Flickr30k (R@10)}} \\
        \midrule
        AltCLIP-18        & 18.5 & 15.0 &  8.0 &  3.7 & 14.8 & 10.1 & 11.5 & 10.2 & 25.2 & 19.5 & 15.6 & 11.7 \\
        SigLIP2           & \underline{55.1} & 52.0 & 25.6 & 20.3 & 37.4 & 34.8 & \underline{44.8} & 43.5 & \underline{49.2} & 45.8 & \underline{42.4}& 39.3 \\
        NLLB-CLIP         & 33.1 & 39.3 &  9.7 & 10.5 & 24.4 & 29.4 & 21.1 & 24.5 & 42.3 & 48.4 & 26.1 & 30.4 \\
        M-CLIP            & 46.6 & \textbf{71.7} & \underline{34.7} & \textbf{63.0} & \underline{43.4} & \textbf{72.0} & 39.7 & \textbf{69.7} & 43.2 & \textbf{71.9} & 41.5 & \textbf{69.7}\\
        \textbf{uCLIP (Ours)} & \textbf{61.5} & \underline{61.3} & \textbf{56.9} & \underline{56.1} & \textbf{60.2} & \underline{59.9} & \textbf{60.4} & \underline{59.6} & \textbf{60.8} & \underline{61.4} & \textbf{60.0}& \underline{59.7}\\
        \midrule
        
        \multicolumn{13}{c}{\textit{XM3600 (R@10)}} \\
        \midrule
        AltCLIP-18        & 23.1 & 20.2 & 12.9 &  8.7 & 22.6 & 14.9 & 17.8 & 16.3 & 35.2 & 27.3 & 22.3 & 17.5 \\
        SigLIP2           & \underline{59.2} & 50.4 & 48.7 & 38.9 & 60.6 & 54.7 & 60.9 & 58.0 & 60.7 & 56.9 & 58.0 & 51.8 \\
        NLLB-CLIP         & 30.8 & 29.9 & 15.8 & 13.9 & 31.2 & 29.5 & 31.2 & 29.5 & 53.7 & 53.2 & 31.7& 30.6\\
        M-CLIP            & 52.1 & \textbf{66.8} & \underline{61.7} & \textbf{74.7} & \underline{66.9} & \textbf{83.9} & \underline{69.1} & \textbf{80.9} & \underline{71.5} & \textbf{81.0} & \underline{64.3}& \textbf{77.5}\\
        \textbf{uCLIP (Ours)} & \textbf{64.5} & \underline{62.8} & \textbf{69.0} & \underline{67.4} & \textbf{77.2} & \underline{76.2} & \textbf{72.7} & \underline{71.6} & \textbf{75.6} & \underline{73.9} & \textbf{71.8}& \underline{70.4}\\
    \bottomrule
    \end{tabular}
    \caption{\textbf{Bidirectional Image–Text retrieval results across five languages.} ``cs, fi, hr, hu, ro" indicates the evaluation language. I→T denotes image-to-text retrieval, and T→I denotes text-to-image retrieval. ``Avg." means average score of the metric to its corresponding method. The table reports retrieval performance on translated MSCOCO, Flickr30k datasets, and existing multilingual dataset XM3600.}
    \label{tab:main_quan}
\end{table*}

\section{Experiments}
\subsection{Implementation details}
We adopt a frozen CLIP\footnote{\texttt{OpenCLIP ViT-B-32-datacomp\_xl\_s13b\_b90k}} model for the vision encoder and use an MPNet-base\footnote{\texttt{paraphrase-multilingual-mpnet-base-v2}} model for the multilingual text encoder, both of which remain frozen during training. We provide further implementation details in Supplementary Material.
\subsection{Multilingual Image–Text Retrieval}
\subsubsection{Setup}
We evaluate our approach on multilingual image-to-text retrieval, text-to-image retrieval, and zero-shot classification tasks across five low-resource languages. We assessed the bidirectional retrieval task on two multilingually translated benchmarks (MSCOCO~\citep{lin2014microsoft} and Flickr30k~\citep{young2014image}) and XM3600~\citep{ThapliyalCrossmodal2022}. For image-to-text retrieval, a query image is embedded using the CLIP image encoder and projected via \( \hat{v}_q = f_C(v_q) \). Multilingual candidate sentences \( {\hat{m}_1, \dots, \hat{m}_K} \) are encoded with the multilingual text encoder and projected via \( f_M(\cdot) \). Cosine similarities \( \text{sim}(\hat{v}_q, \hat{m}_j) \) are computed to retrieve the top-matching captions.

\subsubsection{Results}
Table~\ref{tab:main_quan} reports multilingual bidirectional image–text retrieval results across five low-resource languages (cs, fi, hr, hu, ro) and three benchmarks (MSCOCO, Flickr30k, XM3600). 
On image-to-text retrieval (I→T), \textit{uCLIP} achieves average R@10 scores of 53.2\%, 60.0\%, and 71.8\% on MSCOCO, Flickr30k, and XM3600 respectively, outperforming other baselines. Similarly, for text-to-image retrieval (T→I), it records 53.3\%, 59.7\%, and 70.4\%, achieving the competitive score.
Unlike baselines incorporate extensive multilingual pretraining or direct image-multilingual text or English-multilingual text supervision, \textit{uCLIP} operates without any explicit paired supervision and still achieves superior results---highlighting the effectiveness of its lightweight cross-modal alignment approach, using English as semantic pivot. These results underline the generalizability of \textit{uCLIP}, which achieves competitive multilingual grounding with only 1.7M trainable parameters. Additional Recall@1, Recall@5 scores and qualitative retrieval samples are provided in Supplementary Material.

\subsection{Zero-shot Classification}
\subsubsection{Setup}
We assess zero-shot classification on multilingually translated benchmarks: CIFAR-10~\citep{krizhevsky2009cifar10}, and STL-10~\citep{coates2011stl10}. Each image is encoded via the CLIP encoder and projected using \( f_C(\cdot) \); class names are encoded and projected via \( f_M(\cdot) \). Classification is performed via cosine similarity, and predictions are based on the most similar class. We report average F1 scores to account for class imbalance.
\subsubsection{Results}
\textit{uCLIP} maintains strong class discrimination across languages without using any paired data, unlike baselines such as M-CLIP and SigLIP2, which depend on direct T-T or I-T supervision. For example, in CIFAR-10, \textit{uCLIP} achieves 90.5 in Finnish and 91.9 in Croatian—matching or surpassing M-CLIP and SigLIP2 that rely on translated captions or massive multilingual text-image pairs, respectively. (e.g., 22B data in SigLIP2). While M-CLIP requires paired multilingual-English text for training, \textit{uCLIP} is trained using unpaired datasets. This shows \textit{uCLIP}’s ability to perform effective multilingual zero-shot classification, despite its lightweight design and lack of any direct supervision.
\begin{table}[t]
    \centering
    \setlength{\tabcolsep}{6pt}
    \footnotesize
        \begin{tabular}{l | c   c   c   c  c | c }
        \toprule
        Method &
        cs & fi & hr & hu & ro & {\textbf{Avg.}} \\
        \midrule
        \multicolumn{7}{c}{\textit{CIFAR-10 (F1)}} \\
        \midrule
        AltCLIP-18        & 29.9 &  9.7 & 70.5 & 16.9 & 35.1 & 32.4 \\
        SigLIP2           & \textbf{90.9} & 34.8 & \textbf{91.9} & \textbf{91.6} & \textbf{89.2} & 79.7 \\
        NLLB-CLIP         & 17.2 & 12.1 & 25.4 & 27.8 & 28.4 & 22.2 \\
        M-CLIP            & 88.4 & \underline{74.5} & 87.9 & \underline{89.0} & 85.5 & \underline{85.1} \\
        \textbf{uCLIP (Ours)} & \underline{89.9} & \textbf{90.5} & \textbf{91.9} & 82.3 & \underline{87.0} & \textbf{88.3} \\
        \midrule
        \multicolumn{7}{c}{\textit{STL-10 (F1)}} \\
        \midrule
        AltCLIP-18        & 44.1 & 14.0 & 74.3 & 19.8 & 46.1 & 39.7 \\
        SigLIP2           & \underline{93.4} & 34.4 & \underline{95.5} & \textbf{97.4} & \underline{93.9} & 82.9 \\
        NLLB-CLIP         & 22.2 & 22.6 & 31.5 & 39.1 & 38.0 & 30.7 \\
        M-CLIP            & \textbf{96.6} & \underline{85.9} & \textbf{96.2} & \underline{97.0} & \textbf{95.6} & \textbf{94.3} \\
        \textbf{uCLIP (Ours)} & 88.2 & \textbf{91.8} & 92.3 & 92.8 & 90.3 & \underline{91.1} \\
        \bottomrule
        \end{tabular}
    \caption{\textbf{Zero-shot classification performance across five languages.} The table reports classification performance on translated CIFAR-10 and STL-10 datasets.}
    \label{tab:classification}
\end{table}
\subsection{Efficiency Analysis}
\subsubsection{Train Efficiency}
As summarized in Table~\ref{tab:model_compare}, \textit{uCLIP} is remarkably lightweight, with only 1.7M trainable parameters and no requirement for paired I-T, T-T data. This minimal parameter usage not only reduces memory and compute overhead but also leads to faster convergence during training. 
This high training efficiency makes \textit{uCLIP} particularly suitable for low-resource or constrained computing environments, such as deployment in academic, mobile, or edge scenarios.
\begin{table}[H]
\centering
\setlength{\tabcolsep}{6pt}
\footnotesize
\begin{tabular}{l|c|c|c}
    \toprule
    Method & I-T Pairs & T-T Pairs & Trainable Params \\
    \midrule
    AltCLIP-18 & \checkmark & \checkmark & 563M \\
    SigLIP2   & \checkmark & \checkmark & 375M \\
    NLLB-CLIP  & \checkmark & \checkmark & 501M \\
    M-CLIP     & $\times$ & \checkmark  & 560M \\
    \textbf{uCLIP (Ours)} & $\times$ & $\times$ & \textbf{1.7M} \\
    \bottomrule
\end{tabular}
\caption{\textbf{Comparison of supervision and model scale.} We compare recent multilingual VLMs in terms of their reliance on I-T pairs and T-T pairs during training, as well as the number of trainable parameters. Unlike prior methods, which require extensive supervision and full encoder tuning, \textit{uCLIP} is trained without any paired data and uses only a lightweight 1.7M parameter projection module.}
\label{tab:model_compare}
\end{table}
\begin{table}[H]
    \centering
    \setlength{\tabcolsep}{3pt}
    \footnotesize
    \begin{tabular}{l|c|c|c}
        \toprule
        Method & \begin{tabular}{c}Translation\\(ms/sample)\end{tabular} & \begin{tabular}{c}Encoding\\ (ms/sample)\end{tabular} & \begin{tabular}{c}Total\\(ms/sample)\end{tabular} \\
        \midrule
        CLIP + Translator      & 459.02 & 21.11 & {480.13} \\
        AltCLIP-18             & $\times$ & 60.42 & {60.42} \\
        SigLIP2               & $\times$ & 47.85 & {47.85} \\
        NLLB-CLIP              & $\times$ & 29.41 & \underline{29.41} \\
        M-CLIP                 & $\times$ & 94.4 & {94.4} \\
        \textbf{uCLIP (Ours)} & $\times$ & 23.56 & \textbf{23.56} \\
        \bottomrule
    \end{tabular}
    \caption{\textbf{Inference time comparison.} We report average per-query inference time in milliseconds, including translation and encoding.}
    \label{tab:inference_time}
\end{table}
\subsubsection{Inference-time Efficiency}
Following prior work such as mCLIP~\citep{chen-etal-2023-mclip}, which compares inference latency between translation-based and multilingual approaches, a common baseline for multilingual text–image retrieval is to first translate non-English queries into English and then apply a pre-trained CLIP model. However, as shown in Table~\ref{tab:inference_time}, this strategy incurs substantial latency, requiring 480.13 ms per query. In contrast, \textit{uCLIP} removes the need for translation and achieves a total inference time of just 23.56 ms—over 20 times faster than translation-based methods—while also being more efficient than other multilingual baselines. The reported inference time is the sum of average image encoding and text encoding time per sample, averaged across five languages. It excludes data loading and batching overheads outside the model pipeline. All measurements were conducted on a single NVIDIA TITAN Xp GPU.

\subsection{Embedding Space Analysis}
\subsubsection{Cosine Similarity Alignment}
\begin{figure}[t]
  \centering
  \includegraphics[width=\linewidth]{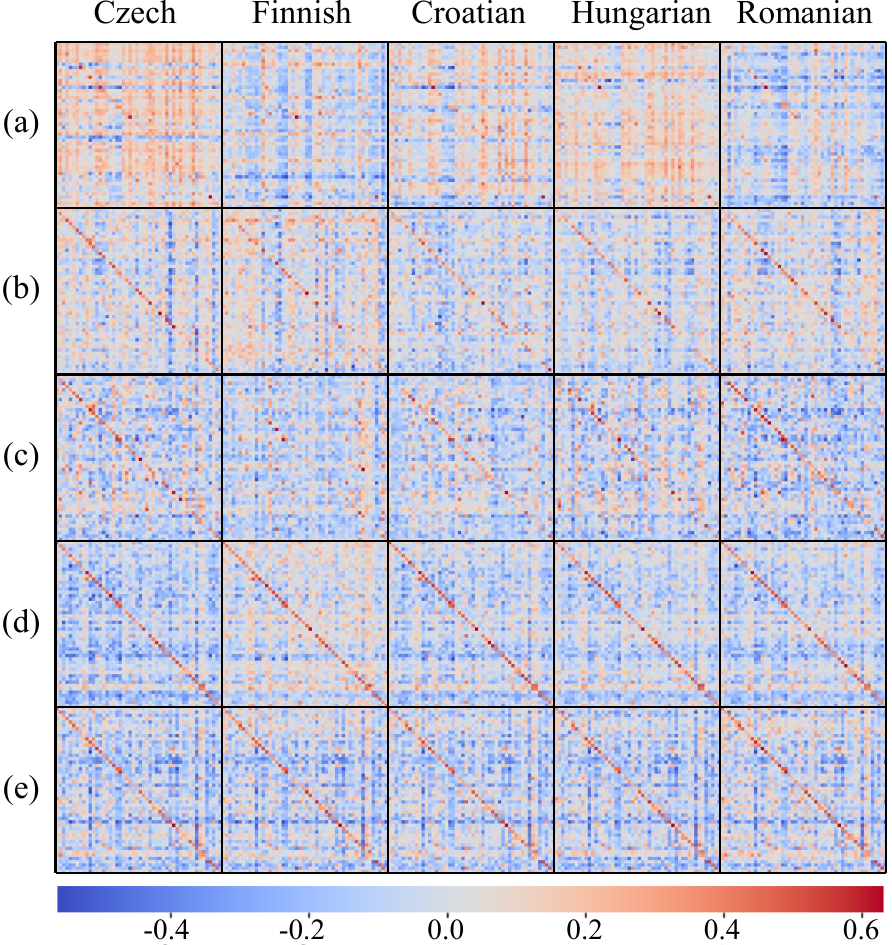}
    \caption{
    \textbf{
    Cosine similarity visualization for embeddings of text and image queries.} We visualize the cosine similarity matrices under five different settings: (a) AltCLIP-18, (b) SigLIP2, (c) NLLB-CLIP, (d) M-CLIP, and (e) using our proposed \textit{uCLIP} model. All text samples are translated in each five language from Flickr30k benchmark.}
  \label{fig:cosim_full}
\end{figure}

To evaluate multilingual text-image alignment, we visualize cosine similarity matrices between 50 image–text pairs from translated Flickr30k under five settings. Figure~\ref{fig:cosim_full} presents the similarity heatmaps for five low-resource languages across various models. A non-scattered strong red diagonal pattern indicates accurate alignment between corresponding image and caption embeddings, while low similarity values (darker blue) in the off-diagonal regions suggest the model correctly suppresses mismatched pairs. Notably, \textit{uCLIP} exhibits both the sharpest diagonal and the darkest off-diagonal regions across all five languages, demonstrating its superior ability to capture correct multilingual alignment while avoiding false positives.
\subsubsection{Visual Representation via UMAP}
To evaluate the robustness of our projection module across different image encoders, we visualize the image embeddings using UMAP on the CIFAR-10 dataset. 
Each point represents an image embedding colored by its ground-truth class. Compared to the raw CLIP and SigLIP2 embeddings, our \textit{uCLIP} projection (b), (d) produces similarly well-separated or even more compact clusters, especially for visually ambiguous classes such as car and truck. Notably, when switching the image encoder from CLIP to SigLIP2, the class separability improves, and our projection (d) preserves this improvement without degradation. These results indicate that the proposed projection module generalizes well across different visual backbones and maintains discriminative structures in the embedding space, even without retraining for each encoder.
\begin{figure}[t]
  \centering
  \includegraphics[width=\linewidth]{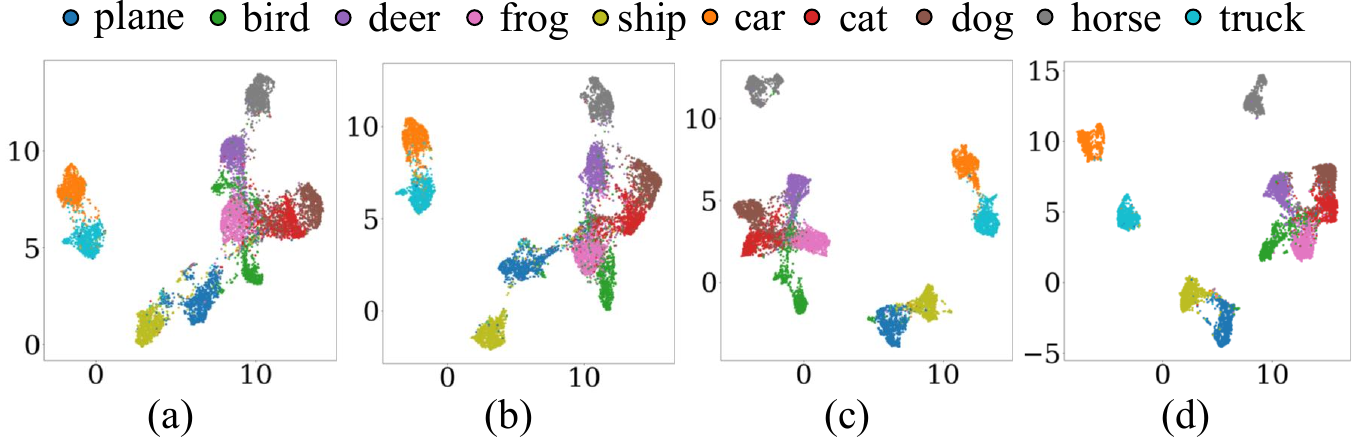}
    \caption{
    \textbf{UMAP visualization of image embedding.} We visualize embeddings extracted from (a) CLIP image encoder, (b) \textit{uCLIP} model with CLIP image encoder, (c) SigLIP2 image encoder, and (d) \textit{uCLIP} model with SigLIP2 image encoder. Results are based on CIFAR-10.}
  \label{fig:umap}
\end{figure}
\begin{table}[t]
    \centering
    \setlength{\tabcolsep}{4pt}
\footnotesize
    \begin{tabular}{ l | c c c | c c c}
        \toprule
        \multirow{2}{*}{\begin{tabular}{c}Multilingual\\Encoder\end{tabular}} &
        \multicolumn{3}{c|}{Image $\rightarrow$ Text} &
        \multicolumn{3}{c}{Text $\rightarrow$ Image} \\
         & R@1 & R@5 & R@10 & R@1 & R@5 & R@10 \\
        \midrule
         E5-base & 9.0   & 23.0     & 32.2     & 8.5   & 23.0     & 31.9     \\
         MiniLM-L12   & 17.1   & 37.6     & 48.9     & 17.6   & 38.8     & 50.6     \\
         \textbf{MPNet-base}   & \textbf{19.9}   & \textbf{41.6}     & \textbf{53.2}     & \textbf{19.3}   & \textbf{41.5}     & \textbf{53.3}     \\  
        \bottomrule
    \end{tabular}
    \caption{\textbf{Effect of multilingual text encoder choice.} We conduct an ablation study comparing different multilingual text encoders. The results are reported on translated MSCOCO dataset.}
    \label{tab:ablation_encoder}
\end{table}
\section{Ablation Studies}
\subsection{Ablation on Encoders}
We evaluate several multilingual text encoders, multilingual versions of E5-base\cite{wang2024multilinguale5textembeddings}\footnote{\texttt{multilingual-e5-base}}, MiniLM-L12\cite{minilm}\footnote{\texttt{paraphrase-multilingual-MiniLM-L12-v2}}, and MPNet-base\cite{Song2020mpnet}\footnote{\texttt{paraphrase-multilingual-mpnet-base-v2}}, which differ in model size and architectural characteristics. For simplicity, we refer to them as E5-base, MiniLM-L12, and MPNet-base throughout the paper. As shown in Table~\ref{tab:ablation_encoder}, MPNet-base achieves the best overall performance, followed by MiniLM-L12 and E5-base. Notably, MiniLM-L12 outperforms E5-base despite having a smaller model size.

To further evaluate the generalizability of our method to inherently multilingual vision-language models, we replace the vision encoder with SigLIP2. The results in Table~\ref{tab:ablation_siglip} demonstrate that our approach enhances the performance of SigLIP2, indicating its effectiveness even when applied to native multilingual VLMs.

\subsection{Ablation on Methodology}
We leverage three loss terms to align the multilingual text encoder with the CLIP vision encoder: $\mathcal{L}_{\text{pseudo}}$, $\mathcal{L}_{\text{text}}$, and $\mathcal{L}_{\text{intra}}$. Note that $\mathcal{L}_{\text{pseudo}}$ and $\mathcal{L}_{\text{text}}$ together constitute the inter-alignment loss, $\mathcal{L}_{\text{inter}}$. We conduct an ablation study to examine the impact of each loss component on performance. The results in Table~\ref{tab:ablation_loss} show that $\mathcal{L}_{\text{intra}}$ and $\mathcal{L}_{\text{text}}$ have a relatively minor effect, while removing $\mathcal{L}_{\text{pseudo}}$ leads to a significant performance drop. This suggests that $\mathcal{L}_{\text{pseudo}}$ serves as the primary mechanism for directly aligning the multilingual text embedding space with the image embedding space, while $\mathcal{L}_{\text{text}}$ and $\mathcal{L}_{\text{intra}}$ play a supportive role by contributing to indirect alignment.  Also, we adopt embedding perturbation to promote local semantic smoothness and improve robustness. As shown in Table~\ref{tab:ablation_loss}, the absence of embedding perturbation substantially degrades performance. This suggests that, without perturbation, pseudo embeddings tend to align too directly with each other, rather than being aligned with the local semantic neighborhoods of their counterparts
\begin{table}[t]
    \centering
    \setlength{\tabcolsep}{3pt}
    \footnotesize
    \begin{tabular}{l | ccc | ccc}
        \toprule
        \multirow{2}{*}{Method} &
        \multicolumn{3}{c|}{Image $\rightarrow$ Text} &
        \multicolumn{3}{c}{Text $\rightarrow$ Image} \\
        & R@1 & R@5 & R@10 & R@1 & R@5 & R@10 \\
        \midrule
        SigLIP2 & 18.1 & 35.5 & 44.7 & 14.5 & 30.5 & 39.4 \\
        \textbf{Ours (w/ SigLIP2)} & \textbf{20.7} & \textbf{42.9} & \textbf{54.6} & \textbf{21.3} & \textbf{43.5} & \textbf{55.1} \\
        \bottomrule
    \end{tabular}
    \caption{\textbf{Effect of our method on SigLIP2.} \textit{uCLIP} significantly improves multilingual retrieval performance over the SigLIP2 baseline. The results are reported on translated MSCOCO dataset.}
    \label{tab:ablation_siglip}
\end{table}
\begin{table}[t]
    \centering
    \setlength{\tabcolsep}{5pt}
    \footnotesize
    \begin{tabular}{l | c c c | c c c}
    \toprule
    \multirow{2}{*}{Method} & \multicolumn{3}{c|}{Image → Text} & \multicolumn{3}{c}{Text → Image} \\
                            & R@1 & R@5 & R@10 & R@1 & R@5 & R@10 \\
    \midrule
        Full & \textbf{19.9} & \textbf{41.6} & \textbf{53.2} & \textbf{19.3} & \textbf{41.5} & \textbf{53.3} \\
    \midrule
    \multicolumn{7}{c}{\textit{Ablations}} \\
    \midrule
    w/o $\mathcal{L}_{\text{intra}}$  & 19.1 & 40.6 & 52.5 & 18.8 & 40.8 & 53.0 \\
    w/o $\mathcal{L}_{\text{text}}$   & 18.1 & 39.8 & 51.2 & 18.2 & 40.0 & 51.9 \\
    w/o $\mathcal{L}_{\text{pseudo}}$ & 16.2 & 36.1 & 47.7 & 18.5 & 40.7 & 52.4 \\
    w/o E.P.    & 9.3   & 23.2 & 32.6 & 16.9 & 37.4 & 48.6 \\
    \bottomrule
    \end{tabular}
    \caption{\textbf{Effect of loss terms and embedding perturbation.} We selectively remove components in our model to evaluate their individual contribution to performance. ``E.P." indicates Embedding Perturbation. The results are reported on translated MSCOCO dataset.}
    \label{tab:ablation_loss}
\end{table}
\section{Conclusion}
\subsection{Summary}
We present \textit{uCLIP}, a parameter-efficient framework that addresses the fundamental challenge of extending vision-language models to underrepresented languages.
Our key contributions include: (1) a novel pivot-based alignment strategy that leverages English representations as semantic anchors to bridge pretrained vision and multilingual text encoders, eliminating the need for paired I-T, T-T datasets during training; and (2) a lightweight projector architecture with only 1.7M trainable parameters—over 99\% smaller than existing multilingual models—while keeping both vision and text encoders completely frozen. The simplicity and effectiveness of our method make it a practical solution for democratizing vision-language capabilities across diverse linguistic communities with minimal computational resources and training overhead.
\subsection{Future Work}
A promising direction is to extend our approach to more languages using stronger multilingual encoders and cross-lingual transfer. We also plan to adapt the model to generation tasks such as captioning and VQA, enabling more complex cross-modal understanding. 
%

\section{Acknowledgments}
This work was supported by the Institute of Information \& Communications Technology Planning \& Evaluation (IITP) grant funded by the Korea government (MSIT) under the Artificial Intelligence Star Fellowship support program to nurture the best talents (IITP-2025-RS-2025-02304828).

\bibliography{aaai2026}
\clearpage
\appendix
\setcounter{secnumdepth}{1}
\setcounter{section}{0}
\setcounter{figure}{0}
\setcounter{table}{0}
\renewcommand{\thesection}{\Alph{section}}
\renewcommand{\thefigure}{S\arabic{figure}}
\renewcommand{\thetable}{S\arabic{table}}

\begin{center}
{\LARGE\bf Appendix\par}
\end{center}
\vspace{1em}


\section{Related work}
\label{appendix:related_work}
\subsection{Vision-Language Pretrained Models}
Vision-Language Pretrained Models (VL-PTMs) can be broadly categorized based on  how they model interactions between image and text modalities. Fusion encoder models jointly process multimodal inputs, either through a single-stream architecture~\citep{Huang_2021_SOHO,li2021unimo,Xia2021XGPT} or dual-stream designs with cross-attention modules~\citep{huo2021wenlanbridgingvisionlanguage,li2021ALBEF,xue2021visualparsing}. In contrast, dual encoder models such as CLIP~\citep{radford2021learning}, ALIGN~\citep{jia2021scaling}, and DeCLIP~\citep{smeu2025declip} encode image and text inputs independently via modality-specific encoders and map them into a shared embedding space. This design allows efficient similarity computation and scalability for large-scale retrieval tasks. Hybrid models~\citep{bao2022VLMo,Singh_2022_FLAVA} have also emerged, combining the strengths of both fusion and dual encoder architectures. However, the success of VL-PTMs fundamentally depends on the availability of large-scale paired image–text data, which remains predominantly English-centric. This reliance poses a major challenge when adapting to languages with limited multimodal resources.

\subsection{Cross-Lingual and Multilingual Extensions}
To expand cross-lingual applicability, several efforts have adapted CLIP-like models to support non-English languages. 
M-CLIP~\citep{carlsson2022multilingualclip} performs text–text knowledge distillation from the CLIP text encoder to mBERT, requiring large-scale parallel corpora for multilingual supervision.
AltCLIP~\citep{chen2022altclip} aligns XLM-R to the CLIP text encoder using text–text pairs and further fine-tunes it with image–text data to align with the frozen CLIP image encoder.
NLLB-CLIP~\citep{visheratin2023nllb} leverages the NLLB~\citep{costa2022no} machine translation model and fine-tunes a multilingual encoder using translated captions paired with CLIP image embeddings, relying on image–text supervision.
While effective, these encoder-replacing approaches depend on some form of paired supervision—either text–text or image–text—to align the multilingual encoder with the visual space.
On the other hand, SigLIP2~\citep{zhai2023siglip} adopts a joint training framework with sigmoid loss, using massive multilingual datasets to achieve strong multilingual capabilities. However, this design incurs significant computational cost and demands extensive multilingual resources. 
Most existing methods still rely heavily on large-scale paired data in the target language—either image–text pairs or parallel corpora—which are often unavailable or extremely limited in low-resource scenarios. This reliance makes it difficult to train or even adapt such models for underrepresented languages, thereby limiting their practical applicability in multilingual settings. These limitations motivate the development of alternative approaches that leverage high-resource pivot languages—such as English—to enable cross-lingual vision-language alignment without requiring direct supervision in the target language.

\subsection{Indirect Cross-Modal Alignment}
Recent works on cross-modal representation learning have proposed lightweight and indirect alignment frameworks that minimize parameter overhead while avoiding direct supervision between modality pairs. For instance, C-MCR~\citep{wang2023connecting} introduces a contrastive objective with shared text anchors to align modality pairs such as image and audio. Similarly, mCLIP~\citep{chen-etal-2023-mclip} distills knowledge from CLIP into a multilingual encoder using lightweight projection heads, thereby reducing the need for full model fine-tuning.

\section{Implementation details}
\label{appendix:implementation}
\subsection{Hyperparameters}
The temperature parameters $\tau$ used in Equations (1), (2) and (5) are all set to \(1/100\). For noise injection in Equations (3) and (4), the Gaussian noise variance \(\sigma^2\) is set to 0.004. We train the projectors for 5 epochs using a batch size of 2048. The AdamW optimizer is used with an initial learning rate of \(1e\text{-}3\), and the learning rate is scheduled with linear decay. All experiments are conducted with frozen encoders, and only the projection layers are updated during training. Training takes approximately 2.5 hours using four NVIDIA A100 40GB GPUs.
\subsection{Architecture of projection layer}
The two projectors, \(f_C(\cdot)\) and \(f_M(\cdot)\), follow similar multi-layer perceptron (MLP) architectures, each consisting of two linear layers with an intermediate BatchNorm1D and ReLU activation, as detailed in Table~\ref{tab:projection_layer}. Specifically, \(f_C(\cdot)\) takes a 512-dimensional input, expands it to 1024, applies BatchNorm1D and ReLU, and then projects it down to 512 dimensions. On the other hand, \(f_M(\cdot)\) starts from a 768-dimensional input, expands to 1536, and similarly applies BatchNorm1D and ReLU before reducing the dimensionality to 512. This consistent structure ensures that both visual and multilingual features are projected into a shared 512-dimensional embedding space while accounting for differences in the input sizes of their respective encoders.
\begin{table}[h]
    \centering
    \setlength{\tabcolsep}{6pt}
    \footnotesize
    \begin{tabular}{l|l|c|c}
        \toprule
        Module & Block & $C_{in}$ & $C_{out}$ \\
        \midrule
        \multirow{4}{*}{Projector $f_C$} 
        & Linear         & 512   & 1024 \\
        & BatchNorm1D    & -  & - \\
        & ReLU           & -     & -    \\
        & Linear         & 1024  & 512 \\
        \midrule
        \multirow{4}{*}{Projector $f_M$} 
        & Linear         & 768   & 1536 \\
        & BatchNorm1D    & -  & - \\
        & ReLU           & -     & -    \\
        & Linear         & 1536  & 512 \\      
        \bottomrule
    \end{tabular}
    \caption{\textbf{Architecture of projection layer} }
    \label{tab:projection_layer}
\end{table}

\subsection{Datasets}
We construct the English text dataset by mixing moondream2-coyo-5M-captions, a refined version of the COYO~\citep{kakaobrain2022coyo-700m} dataset with captions generated by Moondream2~\citep{moon2coyo2024}, and BLIP3o-Pretrain-Short-Caption, which consists of short captions generated by Qwen2.5-VL~\citep{bai2025qwen2}.
For the multilingual text memory, we sample 0.4 million captions for each of the 5 languages (Czech, Swedish, Croatian, Indonesian, and Norwegian) from the laion-coco-nllb  dataset, a translated version of LAION-5B~\cite{schuhmann2022laion}.
The image memory consists of 2 million images, randomly sampled from ImageNet-1K~\citep{ImageNet1K} (42.5\%), COCO~\citep{lin2014microsoft} (5.1\%), and CC12M~\citep{changpinyo2021conceptual} (50.4\%).
Most of the datasets above are accessed via the Huggingface Datasets library~\citep{lhoest2021datasets}.
Note that the image memory, English text dataset, and multilingual text memory are not paired with each other. They can be freely replaced with image-only or text-only data, including AI-generated content.
\section{Additional Results}
\subsection{Retrieval results}
Table~\ref{tab:retrieval_full} presents comprehensive bidirectional image-text retrieval results across three multilingual benchmarks (MSCOCO, Flickr30k, and XM3600), including Recall@1, Recall@5, and Recall@10 for both image-to-text (I→T) and text-to-image (T→I) retrieval tasks. Our method demonstrates consistently strong performance across all three datasets and evaluation metrics, with particularly higher performance in image-to-text retrieval tasks.
When examining the average performance across all evaluated languages, \textit{uCLIP} demonstrates remarkable efficiency relative to its computational requirements. Despite utilizing only 1.7M trainable parameters and requiring no paired multilingual image-text and text-text data during training, our method achieves an average I→T Recall@10 of 61.7\% and T→I Recall@10 of 61.2\% across all benchmarks. This performance is particularly impressive considering that competing methods require substantially larger trainable parameter counts and extensive multilingual supervision, yet \textit{uCLIP} achieves comparable or superior results with minimal computational overhead and training data requirements.

\begin{table*}[t]
    \centering
    \setlength{\tabcolsep}{1.5pt}
    \footnotesize
    \begin{tabular}{l c | ccc | ccc | ccc | ccc | ccc | ccc}
        \toprule
        \multirow{3}{*}{Method} & \multirow{3}{*}{Lang.} &
        \multicolumn{6}{c|}{MSCOCO} &
        \multicolumn{6}{c|}{Flickr30k} &
        \multicolumn{6}{c}{XM3600} \\
        \cmidrule(lr){3-8} \cmidrule(lr){9-14} \cmidrule(lr){15-20}
        & & \multicolumn{3}{c|}{I$\rightarrow$T} & \multicolumn{3}{c|}{T$\rightarrow$I} &
        \multicolumn{3}{c|}{I$\rightarrow$T} & \multicolumn{3}{c|}{T$\rightarrow$I} &
        \multicolumn{3}{c|}{I$\rightarrow$T} & \multicolumn{3}{c}{T$\rightarrow$I} \\
        & & R@1 & R@5 & R@10 & R@1 & R@5 & R@10 &
        R@1 & R@5 & R@10 & R@1 & R@5 & R@10 &
        R@1 & R@5 & R@10 & R@1 & R@5 & R@10 \\
        \midrule
        AltCLIP-18 & cs & 5.6 & 12.9 & 18.2 & 4.5 & 11.4 & 16.0 & 6.6 & 14.0 & 18.5 & 5.3 & 11.1 & 15.0 & 0.9 & 18.5 & 23.1 & 8.0 & 16.0 & 20.2\\
    
    SigLIP2 & cs & \textbf{22.4} & \underline{41.8} & \underline{51.4} & 17.0 & 36.5 & 46.7 & \underline{20.5} & \underline{44.3} & \underline{55.1} & 23.7 & 43.2 & 52.0 & \textbf{27.3} & \underline{50.4} & \underline{59.2} & 21.0 & 41.9 & 50.4 \\
    
    NLLB-CLIP & cs & 8.4 & 20.7 & 28.6 & 9.6 & 23.1 & 31.3 & 11.2 & 25.1 & 33.1 & 14.3 & 30.3 & 39.3  & 10.3 & 23.9 & 30.8 & 10.6 & 23.1 & 29.9 \\
    
    M-CLIP & cs & 9.7 & 27.9 & 40.1 & \textbf{21.4} & \textbf{43.4} & \textbf{54.8} & 12.6 & 34.0 & 46.6 & \textbf{35.7} & \textbf{62.1} & \textbf{71.7} & 17.7 & 39.1 & 52.1 & \textbf{28.9} & \textbf{55.3} & \textbf{66.8} \\
    
    \textbf{uCLIP (Ours)} & cs & \underline{20.7} & \textbf{42.5} & \textbf{54.9} & \underline{19.5} & \underline{42.3} & \underline{54.2} & \textbf{25.7} & \textbf{50.2} & \textbf{61.5} & \underline{26.2} & \underline{50.6} & \underline{61.3} & \underline{27.1} & \textbf{52.6} & \textbf{64.5} & \underline{24.3} & \underline{51.1} & \underline{62.8} \\
    
    \midrule
    AltCLIP-18 & fi & 1.9 & 5.0 & 7.6 & 1.3 & 3.0 & 4.6 & 2.3 & 5.4 & 8.0 & 0.9 & 2.4 & 3.7 & 5.6 & 10.2  & 12.9 & 3.5 & 6.9 & 8.7 \\
    SigLIP2 & fi & \underline{10.8} & \underline{23.2} & 30.6 & 7.3 & 18.0 & 24.6 & \underline{9.0} & 19.7 & 25.6 & 6.0 & 14.6 & 20.3 & 21.9 & 40.6 & 48.7 & 16.1 & 31.1 & 38.9 \\
    NLLB-CLIP & fi & 2.6 & 7.0 & 10.1 & 2.0 & 5.9 & 8.7 & 2.4 & 6.5 & 9.7 & 2.6 & 7.2 & 10.5 & 5.1 & 11.9 & 15.8 & 4.2 & 10.4 & 13.9 \\
    M-CLIP & fi & 7.5 & 21.0 & \underline{31.8} & \underline{17.4} & \underline{37.8} & \underline{48.1} & 7.4 & \underline{24.7} & \underline{34.7} & \textbf{27.7} & \textbf{52.8} & \textbf{63.0} & \underline{22.8} & \underline{49.0}  & \underline{61.7} & \textbf{37.7} & \textbf{65.1} & \textbf{74.7} \\
    
    \textbf{uCLIP (Ours)} & fi & \textbf{17.7} & \textbf{38.9} & \textbf{50.0} & \textbf{17.8} & \textbf{38.6} & \textbf{49.9} & \textbf{22.9} & \textbf{45.9} & \textbf{56.9} & \underline{23.3} & \underline{45.8} & \underline{56.1} & \textbf{31.2} & \textbf{58.2} & \textbf{69.0} & \underline{30.3} & \underline{55.8} & \underline{67.4} \\
    
    \midrule
    AltCLIP-18 & hr & 4.3 & 10.0 & 14.6 & 2.7 & 7.8 & 11.8 & 5.1 & 11.2 & 14.8 & 3.2 & 7.3 & 10.1 & 8.8 & 17.3 & 22.6 & 5.9 & 12.1 & 14.9  \\
    SigLIP2 & hr & \underline{14.7} & \underline{31.1} & 40.0 & 11.9 & 26.3 & 35.2 & \underline{14.9} & 29.2 & 37.4 & 13.2 & 27.2 & 34.8 & \underline{28.1} & 51.6 & 60.6 & 26.0 & 45.7 & 54.7 \\
    NLLB-CLIP & hr & 7.0 & 16.6 & 23.0 & 7.0 & 18.1 & 25.1 & 8.2 & 18.5 & 24.4 & 9.8 & 22.2 & 29.4 & 11.6 & 23.7 & 31.2 & 10.9 & 22.8 & 29.5 \\
    M-CLIP & hr & 10.7 & 30.4 & \underline{42.7} & \textbf{20.9} & \textbf{43.4} & \textbf{54.8} & 10.7 & \underline{30.8} & \underline{43.4} & \textbf{34.9} & \textbf{61.8} & \textbf{72.0} & 21.8 & \underline{53.3} & \underline{66.9} & \textbf{47.0} & \textbf{74.7} & \textbf{83.9} \\
    
    \textbf{uCLIP (Ours)} & hr & \textbf{20.2} & \textbf{42.8} & \textbf{53.6} & \underline{19.9} & \underline{42.4} & \underline{54.3} & \textbf{24.9} & \textbf{49.5} & \textbf{60.2} & \underline{25.4} & \underline{49.1} & \underline{59.9} & \textbf{38.1} & \textbf{66.5} & \textbf{77.2} & \underline{37.6} & \underline{66.3} & \underline{76.2} \\
    
    \midrule
    AltCLIP-18 & hu & 3.2 & 8.0 & 11.5 & 2.7 & 6.8 & 9.6 & 4.3 & 9.0 & 11.5 & 3.4 & 7.6 & 10.2 & 7.7 & 14.3 & 17.8 & 6.8 & 13.0 & 16.3 \\
    SigLIP2 & hu & \underline{20.1} & \underline{38.1} & \underline{48.4} & 16.5 & 34.4 & 43.3 & \underline{19.8} & \underline{37.0} & \underline{44.8} & 17.9 & 35.5 & 43.5 & \underline{30.4} & 51.7 & 60.9 & 26.8 & 48.5 & 58.0 \\
    
    NLLB-CLIP & hu & 6.5 & 16.1 & 22.4 & 6.3 & 16.3 & 23.2 & 6.2 & 14.9 & 21.1 & 7.2 & 18.0 & 24.5 & 9.4 & 20.5 & 27.0 & 9.3 & 19.8 & 26.4 \\
    
    M-CLIP & hu & 9.6 & 28.8 & 42.2 & \textbf{20.5} & \textbf{42.1} & \textbf{53.5} & 9.0 & 28.4 & 39.7 & \textbf{34.1} & \textbf{60.0} & \textbf{69.7} & 27.5 & \underline{57.1} & \underline{69.1} & \textbf{45.0} & \textbf{71.8} & \textbf{80.9} \\
    
    \textbf{uCLIP (Ours)} & hu & \textbf{19.9} & \textbf{41.0} & \textbf{52.2} & \underline{18.9} & \underline{41.4} & \underline{53.2} & \textbf{24.7} & \textbf{48.9} & \textbf{60.4} & \underline{24.8} & \underline{49.0} & \underline{59.6} & \textbf{34.0} & \textbf{62.4} & \textbf{72.7} & \underline{34.5} & \underline{60.7} & \underline{71.6} \\
    
    \midrule
    AltCLIP-18 & ro & 19.9 & 15.5 & 21.6 & 5.1 & 12.8 & 18.4 & 9.6 & 19.1 & 25.2 & 6.4 & 15.0 & 19.5 & 14.3 & 28.6 & 35.2 & 10.8 & 21.8 & 27.3  \\
    SigLIP2 & ro & \textbf{22.6} & \textbf{43.4} & \underline{53.3} & 19.0 & 37.2 & 47.2 & \underline{22.0} & \underline{40.7} & \underline{49.2} & 19.7 & 37.4 & 45.8 & \underline{32.0} & 53.1 & 60.7 & 28.0 & 48.7 & 56.9  \\
    
    NLLB-CLIP & ro & 12.5 & 30.2 & 40.1 & 14.6 & 32.8 & 42.9 & 15.0 & 33.0 & 42.3 & 18.8 & 37.9 & 48.4 & 21.7 & 43.4 & 53.7 & 22.8 & 43.5 & 53.2  \\
    
    M-CLIP & ro & 9.9 & 27.2 & 39.5 & \textbf{22.4} & \textbf{44.7} & \textbf{56.3} & 10.6 & 30.7 & 43.2 & \textbf{35.7} & \textbf{62.3} & \textbf{71.9} & 28.7 & \underline{59.9} & \underline{71.5} & \textbf{44.0} & \textbf{71.4} & \textbf{81.0} \\
    
    \textbf{uCLIP (Ours)} & ro & \underline{20.7} & \underline{42.6} & \textbf{55.4} & \underline{20.4} & \underline{42.7} & \underline{54.8} & \textbf{24.5} & \textbf{49.4} & \textbf{60.8} & \underline{25.7} & \underline{49.8} & \underline{61.4} & \textbf{35.9} & \textbf{64.9} & \textbf{75.6} & \underline{34.1} & \underline{62.4} & \underline{73.9} \\
    
    \midrule\midrule
    AltCLIP-18 & Avg. & 7.0 & 10.3 & 14.7 & 3.3 & 8.4 & 12.1 & 5.6 & 11.7 & 15.6 & 3.8 & 8.7 & 11.7 & 7.5 & 17.8 & 22.3 & 7.0 & 14.0 & 17.5  \\
    SigLIP2 & Avg. & \underline{18.1} & \underline{35.5} & \underline{44.7} & 14.5 & 30.5 & 39.4 & \underline{17.2} & \underline{34.2} & \underline{42.4} & 16.1 & 31.6 & 39.3 & \underline{27.9} & 49.5 & 58.0 & 23.6 & 43.2 & 51.8 \\
    NLLB-CLIP & Avg. & 7.4 & 18.1 & 24.8 & 7.9 & 19.2 & 26.2 & 8.6 & 19.6 & 26.1 & 10.5 & 23.1 & 30.4 & 11.6 & 24.7 & 31.7 & 11.6 & 23.9 & 30.6 \\
    M-CLIP & Avg. & 9.5 & 27.1 & 39.3 & \textbf{20.5} & \textbf{42.3} & \textbf{53.5} & 10.1 & 29.7 & 41.5 & \textbf{33.6} & \textbf{59.8} & \textbf{69.7} & 23.7 & \underline{51.7} & \underline{64.3} & \textbf{40.5} & \textbf{67.6} & \textbf{77.5} \\
    \textbf{uCLIP (Ours)} & Avg. & \textbf{19.8} & \textbf{41.6} & \textbf{53.2} & \underline{19.3} & \underline{41.5} & \underline{53.3} & \textbf{24.5} & \textbf{48.8} & \textbf{60.0} & \underline{25.1} & \underline{48.9} & \underline{59.7}  & \textbf{33.3} & \textbf{60.9} & \textbf{71.8} & \underline{32.2} & \underline{59.3} & \underline{70.4} \\
        \bottomrule
    \end{tabular}
    \caption{\textbf{Bidirectional Image–Text retrieval results.} ``Lang." indicates the evaluation language. I→T denotes image-to-text retrieval, and T→I denotes text-to-image retrieval. ``Avg." means average score of the metric to its corresponding method.}
    \label{tab:retrieval_full}
\end{table*}

\subsection{Qualitative results}
\label{appendix:qualitative_result}
\begin{figure*}[t]
  \centering
    \includegraphics[width=\textwidth]{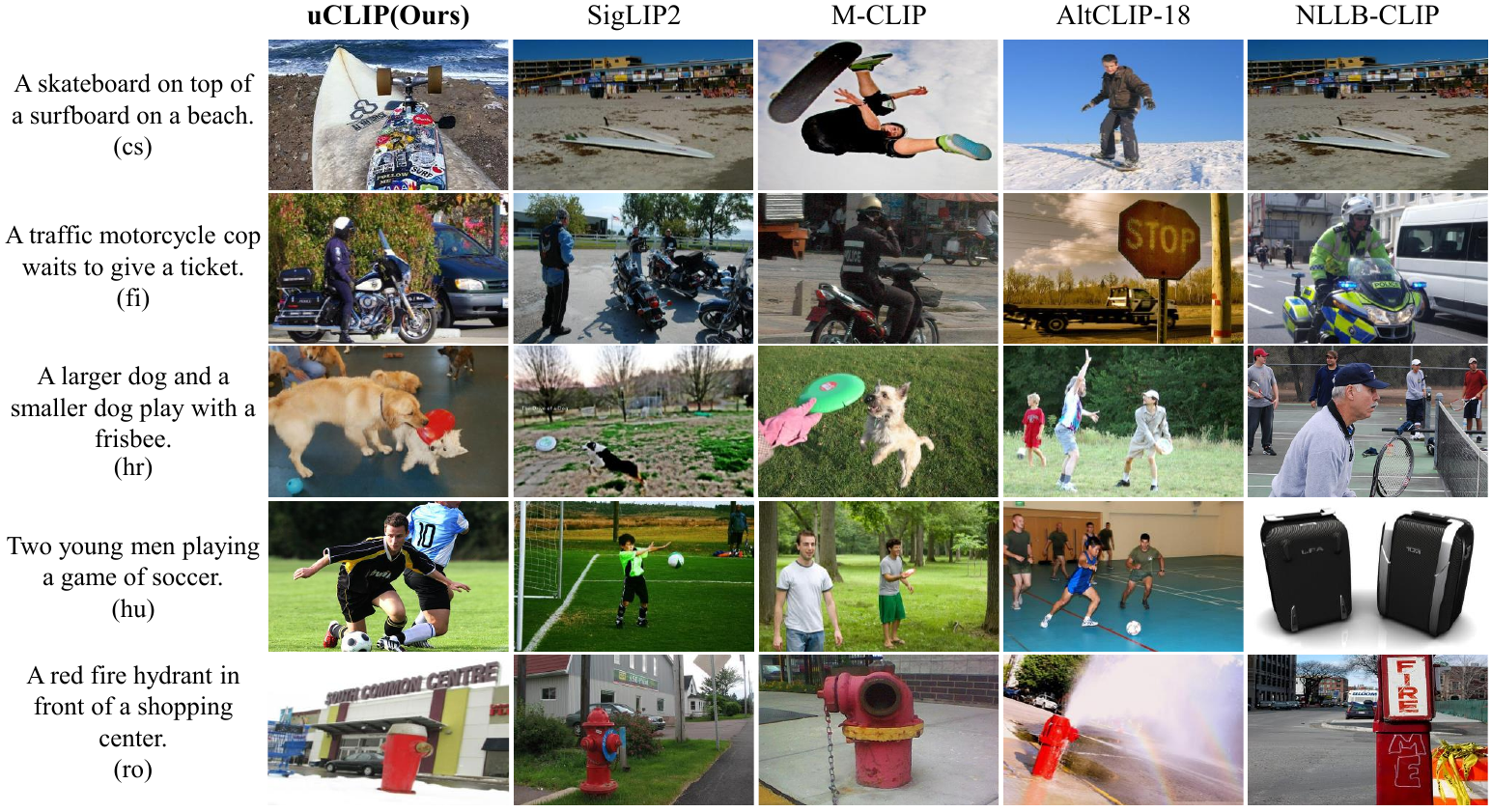}
    \caption{\textbf{Qualitative retrieval results for each language query.}
    We conduct retrieval experiments on the translated MSCOCO dataset, using multilingual text queries in five languages: Czech (cs), Finnish (fi), Croatian (hr), Hungarian (hu), and Romanian (ro). The figure shows top-1 retrieved images for each queries. \textit{uCLIP} consistently retrieves ground-truth-aligned images, while comparison models often fail to retrieve semantically correct results, highlighting the effectiveness of our multilingual alignment strategy. }
\label{retreival_qual}
\end{figure*}

We present qualitative comparisons of multilingual image-text retrieval in Figure~\ref{retreival_qual}. Each query is a translated sentence from one of five low-resource languages, and we compare the top-1 retrieved images from \textit{uCLIP} and four baselines. Compared to existing models, \textit{uCLIP} consistently retrieves more semantically aligned images, demonstrating robust comprehension of not only key objects but also relational and contextual elements in the sentence. 
For example, given the query “Two young men playing a game of soccer,” other models often retrieve images with only one person, scenes with more than two individuals, or even incorrect activities (e.g., frisbee), suggesting failure to capture compositional structure. In contrast, \textit{uCLIP} retrieves a correctly grounded image matching both the subject count and activity. Similarly, for the query “A red fire hydrant in front of a shopping center,” most models detect the hydrant but miss the “shopping center” context, indicating over-reliance on salient object keywords. \textit{uCLIP}, however, correctly captures both the foreground and background, illustrating strong visual grounding and full-sentence understanding.
These results suggest that \textit{uCLIP} is able to go beyond surface-level keyword matching and instead align multimodal features based on deeper cross-lingual and cross-modal semantic structure

\section{Underrepresented Language Selection}
\label{appendix:baselines}
\begin{figure*}[t]
  \centering
  \includegraphics[width=\textwidth]{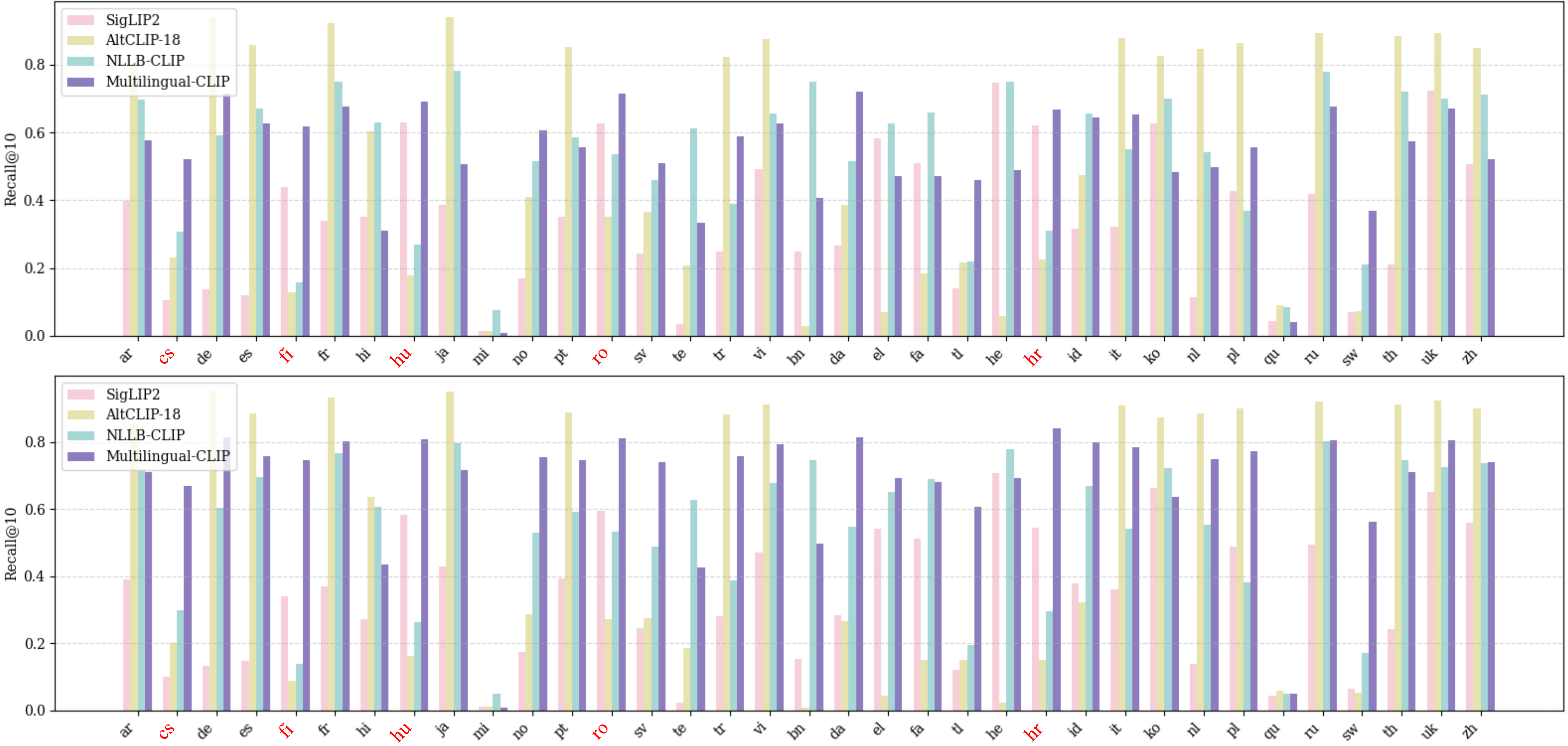}
  \caption{Image-to-text (first row) and text-to-image (second row) Recall@10 score of SigLIP2, AltCLIP-18, NLLB-CLIP, M-CLIP on XM3600 benchmark. Languages in red represent the five low-resource languages we target (cs, fi, hr, hu, ro), selected because four baseline models consistently exhibit low retrieval scores, compared to other languages.}
  \label{fig:xm3600_baselines}
\end{figure*}

\begin{figure*}[t]
  \centering
  \includegraphics[width=\textwidth]{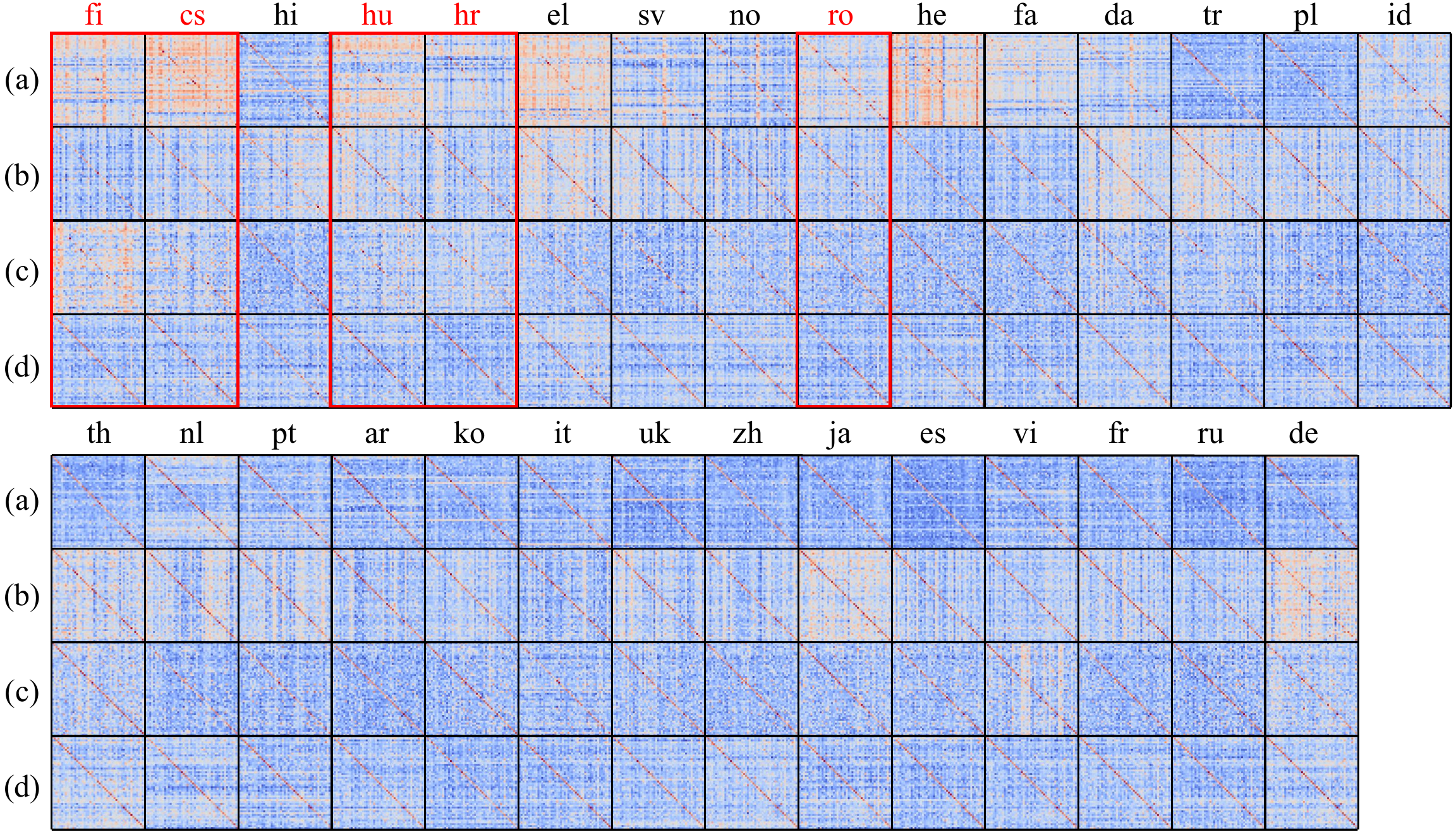}
\caption{
  \textbf{Cosine similarity visualization for embeddings of text and image queries.}
  To identify and analyze languages where existing multilingual vision-language models (VLMs) underperform, we visualize the cosine similarity matrices under four different settings: (a) AltCLIP-18, (b) SigLIP2, (c) NLLB-CLIP, and (d) M-CLIP.
  Each matrix shows pairwise cosine similarities between 50 image-text pairs randomly selected from the XM3600 benchmark.
  The horizontal axis corresponds to image indices, while the vertical axis corresponds to text indices.
  Languages highlighted in red represent the five low-resource languages we target (cs, fi, hr, hu, ro).
  Unsupported languages by our multilingual text encoder are excluded from our evaluation.
}
  \label{fig:cosim_xm3600}
\end{figure*}

To investigate limitations of existing multilingual vision-language models (VLMs), we focus on languages with consistently poor retrieval performance. Figure~\ref{fig:xm3600_baselines} presents Recall@10 scores for image-to-text and text-to-image retrieval tasks across 36 languages on the XM3600 benchmark. Five languages—Czech (cs), Finnish (fi), Croatian (hr), Hungarian (hu), and Romanian (ro)—consistently underperform across all four baseline models. To validate this selection, we visualize cosine similarity matrices of 50 randomly sampled image-text pairs in Figure~\ref{fig:cosim_xm3600}. Strong alignment is indicated by sharp red diagonals and dark blue off-diagonals, while the five selected languages show blurry diagonals and noisy backgrounds, revealing degraded alignment and poor distinction between correct and incorrect pairs. This pattern justifies their designation as underrepresented languages in our study.

\clearpage

\end{document}